\documentclass[letterpaper, 10 pt, conference]{ieeeconf}  % Comment this line out if you need a4paper

\IEEEoverridecommandlockouts                              % This command is only needed if 
\usepackage{graphics} % for pdf, bitmapped graphics files
\usepackage{epsfig} % for postscript graphics files
\usepackage{mathptmx} % assumes new font selection scheme installed
\usepackage{times} % assumes new font selection scheme installed

\usepackage{cite}
\usepackage{amsmath,amssymb,amsfonts}

\usepackage{subfigure}

\usepackage [latin1]{inputenc}

\makeatletter
\title{\LARGE \bf
Generation and Analysis of Feature-Dependent Pseudo Noise for Training Deep Neural Networks
}

\author{Sree Ram Kamabattula$^{1}$ \and Kumudha Musini$^{1}$ \and Babak Namazi$^{2}$ \and Ganesh Sankaranarayanan$^{2}$ \and Venkat Devarajan$^{1}$% <-this % stops a space
\thanks{Correspondence to {\tt\small sreeram.kamabattula@mavs.uta.edu}}
\thanks{$^{1}$University of Texas at Arlington, Electrical/ Biomedical Engineering, Arlington, Texas, USA.}%
\thanks{$^{2}$UT Southwestern Medical Center, Dept. of Surgery, Dallas, Texas, USA.}%
\thanks{\copyright~2021 IEEE. Personal use of this material is permitted. Permission from IEEE must be obtained for all other uses, in any current or future media, including reprinting/republishing this material for advertising or promotional purposes, creating new collective works, for resale or redistribution to servers or lists, or reuse of any copyrighted component of this work in other works.}.
}

\begin{document}

\maketitle
\thispagestyle{empty}
\pagestyle{empty}

%%%%%%%%%%%%%%%%%%%%%%%%%%%%%%%%%%%%%%%%%%%%%%%%%%%%%%%%%%%%%%%%%%%%%%%%%%%%%%%%
\begin{abstract}

Training Deep neural networks (DNNs) on noisy labeled datasets is a challenging problem, because learning on mislabeled examples deteriorates the performance of the network. As the ground truth availability is limited with real-world noisy datasets, previous papers created synthetic noisy datasets by randomly modifying the labels of training examples of clean datasets. However, no final conclusions can be derived by just using this random noise, since it excludes feature-dependent noise. Thus, it is imperative to generate feature-dependent noisy datasets that additionally provide ground truth. Therefore, we propose an intuitive approach to creating feature-dependent noisy datasets by utilizing the training predictions of DNNs on clean datasets that also retain true label information. We refer to these datasets as "Pseudo Noisy datasets". We conduct several experiments to establish that Pseudo noisy datasets resemble feature-dependent noisy datasets across different conditions. We further randomly generate synthetic noisy datasets with the same noise distribution as that of Pseudo noise (referred as "Randomized Noise") to empirically show that i) learning is easier with feature-dependent label noise compared to random noise, ii) irrespective of noise distribution, Pseudo noisy datasets mimic feature-dependent label noise and iii) current training methods are not generalizable to feature-dependent label noise. Therefore, we believe that Pseudo noisy datasets will be quite helpful to study and develop robust training methods.

\end{abstract}

%%%%%%%%%%%%%%%%%%%%%%%%%%%%%%%%%%%%%%%%%%%%%%%%%%%%%%%%%%%%%%%%%%%%%%%%%%%%%%%%
\section{Introduction}
\label{sec:introduction}

Deep neural networks (DNNs) have demonstrated excellent performance in several tasks such as image classification \cite{Krizhevsky2012ImageNetNetworks}, object detection \cite{Redmon2016YouDetection} and several others, owing to the increasing size of training datasets \cite{Deng2010ImageNet:Database} and advanced architectures \cite{He2016DeepRecognition}. However, the labeling process of the large datasets results in a large number of mislabeled examples in several application fields such as medical imaging \cite{Karimi2020DeepAnalysis}, generative networks \cite{Kaneko2019Label-noiseNetworks}, etc. As a result, the generalization error increases, as the DNN learns on the mislabeled training examples \cite{Zhang2017UnderstandingGeneralization}. Therefore, reducing the generalization error while training DNNs with noisy labeled datasets has become prominent work \cite {Algan2019ImageSurvey, Xiao2015LearningClassification}. Mislabeled examples are referred to as noise or label noise throughout this paper.

It is common practice to understand the learning behavior of clean and noisy examples using ground truth, i.e., true labels of training examples \cite{Ma2018Dimensionality-DrivenLabels, Sun2019LimitedLabels, Pleiss2020IdentifyingRanking, Kamabattula2020IdentifyingData}. But unfortunately, most of the real-world noisy datasets do not provide ground truth \cite{Algan2020LabelLearning}. So, previous papers created synthetic noisy datasets by \emph{randomly} changing the true labels of some training examples in clean datasets with distributions such as symmetric and asymmetric while preserving ground truth \cite{Han2018Co-teaching:Labels, Patrini2017MakingApproach, Wang2019SymmetricLabels}.

Several observations and training methods have emerged by exploiting these synthetic noisy datasets. However, \cite{Jiang2020BeyondLabels} pointed out a few contradictory findings and suggested that these methods might not be generalizable to realistic noisy datasets. For example, one such finding is that the generalization error increases, as the DNNs fit the noisy examples \cite{Zhang2017UnderstandingGeneralization}. On the contrary, \cite{Rolnick2017DeepNoise} shows that the DNNs can perform well even in the presence of label noise, when a slightly different noise distribution is adopted. Similarly, another popular small-loss observation becomes less effective with varying noise distributions \cite{Song2019SELFIE:Learning}.

On the other hand, based on these contradictory findings, \cite{Algan2020LabelLearning} suggests that the label noise problem should be treated as three different sub problems, as each type has different characteristics. Specifically, in the order of increasing complexity, they are symmetric (random mislabeling), asymmetric (class-dependent) and feature-dependent (realistic) label noises.

Most of the current research is limited to symmetric and asymmetric noisy datasets due to the lack of ground truth with real world noisy datasets. Consequently, it is imperative to generate feature-dependent noisy datasets that additionally provide ground truth to study and develop robust training methods \cite{Song2020LearningSurvey}.

\subsection{Analyzing learning behavior of label noise}

DNNs learn easier patterns first and fit harder patterns in the later stages of training \cite{Arpit2017ANetworks}. We utilize this fact to distinguish among the three noise categories (symmetric, asymmetric and feature-dependent noise) and, more importantly to understand the learning behavior of feature-dependent label noise.

The noise is uniformly distributed among all the classes in symmetric noise, resembling the case of labelers randomly annotating the dataset \cite{Khetan2017LearningData}. Thus, the network needs to learn harder patterns to fit the \emph{randomly} mislabeled examples. Since, the network fits on harder patterns in the later stages of training, easier patterns mostly refer to clean examples. Consequently, distinguishing between the clean and noisy examples with learning behavior of DNNs is easier with symmetric noise.

Distinguishing between the clean and noisy examples becomes harder with asymmetric noise compared to symmetric noise, as the noise is randomly distributed with just one class in an attempt to mimic the structure of realistic label noise. This can be related to real world noise caused by inexperienced labelers annotating the dataset, where only some randomness is introduced. For example, Red-MiniImageNet \cite{Jiang2020BeyondLabels} real world noisy dataset belongs to this category. The authors have collected $161105$ images by entering the $100$ class names of Mini-ImageNet dataset in a Google search, and verified the labels by multiple human annotators. The verification process showed that there are $54,400$ images with incorrect labels. Now, to create a $\tau$\% noisy training dataset, $\tau$\% of the original training images in the Mini-ImageNet dataset are \emph{randomly} replaced by web images with incorrect labels. For testing, $5000$ images in the ILSVRC12 validation set are used.

\begin{figure}[t]
  
  \centering
 
  \includegraphics[width= 0.255\textwidth]{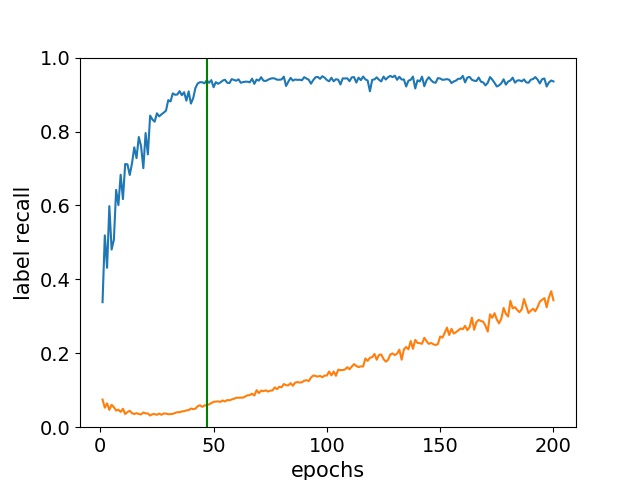}\includegraphics[width= 0.255\textwidth]{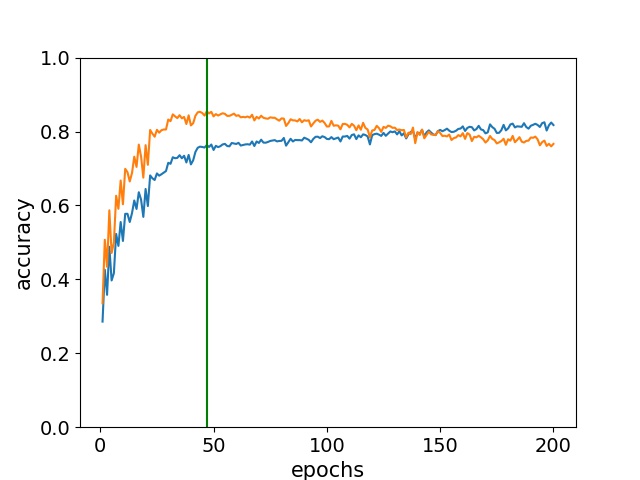}\\
  \includegraphics[width=0.255\textwidth]{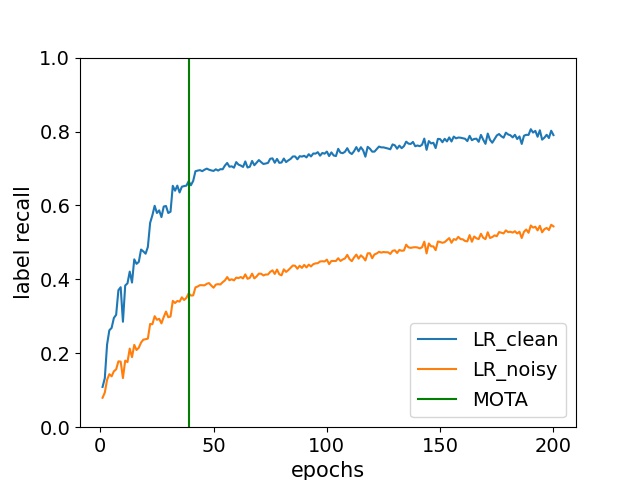}\includegraphics[width= 0.255\textwidth]{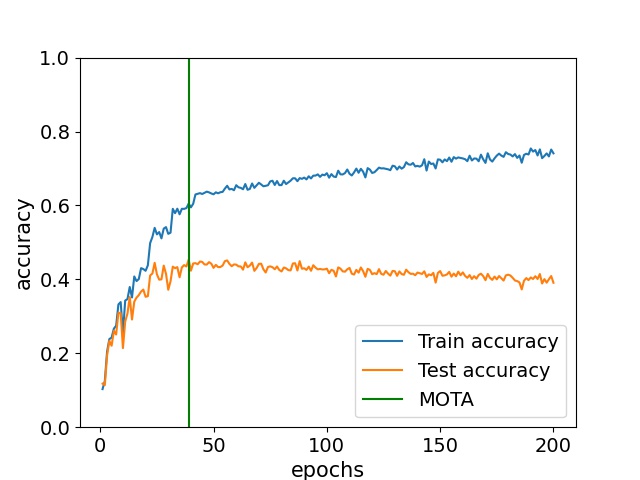}\\
  
  \caption{Label recall (left), Accuracy (right) for symmetric on CIFAR-10 (top), Red-Mini-ImageNet (bottom) for $\tau$ = 0.2 with ResNet32.}
  \label{fig: redmini}
  \vspace{-1\baselineskip}
\end{figure}

We plot the label recall (left) and accuracy (right) of symmetric noise and Red-MiniImageNet noisy dataset in Fig. \ref{fig: redmini}. The experimental details are provided in section \ref{sec: experiments}. With symmetric noise (top plots), it can be observed in the top left plot that the network learns on clean examples $LR_{clean}$ (blue line) at a higher rate initially, while learning on noisy examples $LR_{noisy}$ (orange line) remains minimal, obtaining the maximum test accuracy \emph{MOTA} (vertical green line) in the initial stages of training. As the $LR_{noisy}$ increases in the later stages of training, the test accuracy in the top right plot (orange line) keeps dropping after MOTA.

For Red-MiniImageNet noisy dataset, it can be observed in the bottom left plot that the network learns on both clean and noisy examples from the beginning of the training, unlike the symmetric noise. However, the rate of learning on noisy examples (orange line) is much lower compared to clean examples. As a result, the test accuracy keeps dropping after MOTA, as the train accuracy increases.

\begin{figure}[t]

  \centering
  \includegraphics[width= 0.255\textwidth]{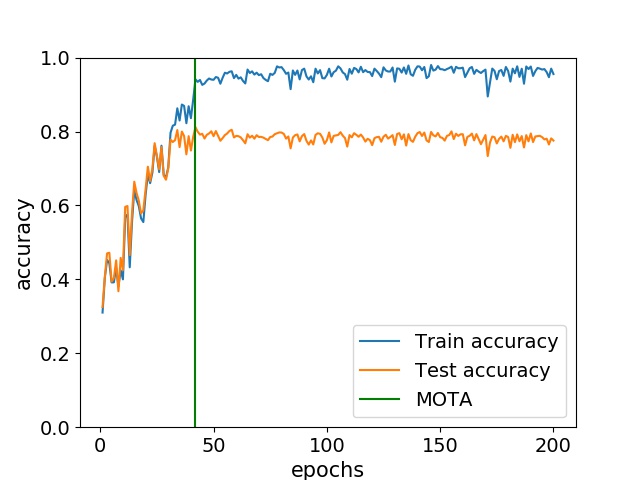}\\
  \caption{Accuracy on ANIMAL-10N with CNN9.}
  \label{fig: animal10n}
  \vspace{-1\baselineskip}

\end{figure}

On the other hand, we noticed in Fig. \ref{fig: animal10n}, a different test accuracy behavior with ANIMAL-10N \cite{Song2019SELFIE:Learning} real world noisy dataset, where the mislabeling happens only between confusing classes. It can be observed that both train and test accuracy are nearly constant in the later stages of training. So, the MOTA doesn't have much significance, since the test accuracy at the end of the training is also the same as that of test accuracy at MOTA. Unfortunately, the learning behavior of clean and noisy examples cannot be monitored, as the ground truth is not provided. However, based on the learning behavior observed in Fig. \ref{fig: redmini}, we believe that this test accuracy behavior results when the network simultaneously learns on clean and noisy examples at a higher rate from the beginning of the training.

With feature-dependent label noise, as the mislabeled examples occur due to similarities between the features, \emph{the network can easily fit both clean and noisy examples from the beginning of training}. In other words, easier patterns refer to both clean and noisy examples with feature-dependent label noise. Thus, distinguishing between clean and noisy examples based on learning behavior of DNNs is much harder. This analysis suggests the hypothesis that the network should easily fit the training examples with feature-dependent label noise compared to random noise.

\subsection{Overview of our work}

In this work, we establish a new class of noisy datasets called \emph{Pseudo} noisy datasets. We refer to these as Pseudo noisy datasets because, we utilize the predictions of a DNN on the training examples of any clean dataset to create noisy datasets of desired noise rates. Thus, Pseudo noisy datasets can render true label information similar to the synthetic noisy datasets. However, it is important to emphasize that, in synthetic noisy datasets, a noise distribution is first chosen to randomly generate noisy labels. On the other hand, with Pseudo noisy datasets, noisy labels in fact arise due to an underlying learning misconception of the DNN. Thus, we claim that the Pseudo noisy datasets would be closer to feature-dependent label noise. We conduct several experiments to show that our created Pseudo noisy datasets resemble feature-dependent noisy datasets.

We further randomly generate synthetic noisy datasets by obtaining the noise distributions from Pseudo noisy datasets to fairly compare random and feature-dependent label noise. For convenience, we call the synthetic noisy datasets with the same noise distribution as that of Pseudo noisy datasets as \emph{Randomized} noisy datasets.

We employ the learning behavior of DNNs on Randomized and Pseudo noisy datasets to i) prove our hypothesis that learning is easier with feature-dependent label noise compared to random noise and ii) prove that Pseudo noisy datasets imitate the feature-dependent label noise irrespective of the noise distribution.

Finally, we empirically show that the existing noise-robust training methods are not generalizable to feature-dependent label noise. Thus, we believe that Pseudo noisy datasets will be quite helpful in developing effective training methods.

\section{Related work}

Our paper focuses on label noise, especially Pseudo noise. To the best of our knowledge, pseudo labels have been used for training in the past \cite{Ding2018ALabels, Reed2015TrainingBootstrapping,Tanaka2018JointLabels}, but they have never been utilized to create feature-dependent noisy datasets.

Web label noise and synthetic noise are closely related to our paper. In web noisy datasets, labeling is often performed by crowd sourcing \cite{Yan2014LearningExpertise}, online queries \cite{Blum2003Noise-tolerantModel}, etc. For example, several clothing images are gathered from online shopping websites and labels are automatically assigned from surrounding texts in Clothing1M noisy dataset \cite{Xiao2015LearningClassification}. A few other widely used web noisy datasets are ANIMAL-10N\cite{Song2019SELFIE:Learning}, Food101N\cite{Lee2018CleanNet:Noise} and WebVision\cite{Li2017WebVisionData}. However, these noisy datasets either do not provide ground truth or only a small clean validation set is available. Therefore, synthetic noisy datasets are exploited to develop different training methods.

One common approach is to select clean samples and train the network on the selected samples \cite{Malach2017DecouplingUpdate, Jiang2018MentorNet:Labels, Chang2017ActiveSamples}. A few methods correct the loss function based on the noise estimation \cite{Patrini2017MakingApproach} and \cite{Goldberger2017TRAININGLAYER}, and assigning higher weights to clean samples \cite{Ren2018LearningLearning} and \cite{Lee2018CleanNet:Noise}. Some research focuses on i) developing noise-robust loss functions \cite{Wang2019SymmetricLabels}, \cite{Zhang2018GeneralizedLabels} and \cite{Wang2019IMAEMatters}, ii) identifying an early training stop point \cite{Kamabattula2020IdentifyingData,Ma2018Dimensionality-DrivenLabels} and \cite{Song2019Prestopping:Noise} and iii) semi-supervised learning \cite{Jiang2020BeyondLabels} and \cite{Li2020DIVIDEMIX:LEARNING}. However, the conclusions cannot be derived using these synthetic noisy datasets as they are just random noise \cite{Jiang2020BeyondLabels}. Therefore, \cite{Algan2020LabelLearning, Xia2020Parts-dependentNoise} and \cite{Cheng2020LearningApproach} developed label-corruption algorithms to synthetically create feature dependent noisy datasets.

\begin{figure}[b]
  \centering
  \includegraphics[width=0.255\textwidth]{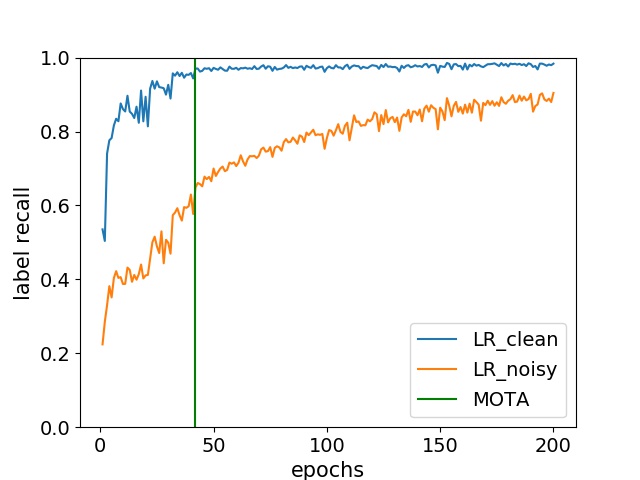}\includegraphics[width= 0.255\textwidth]{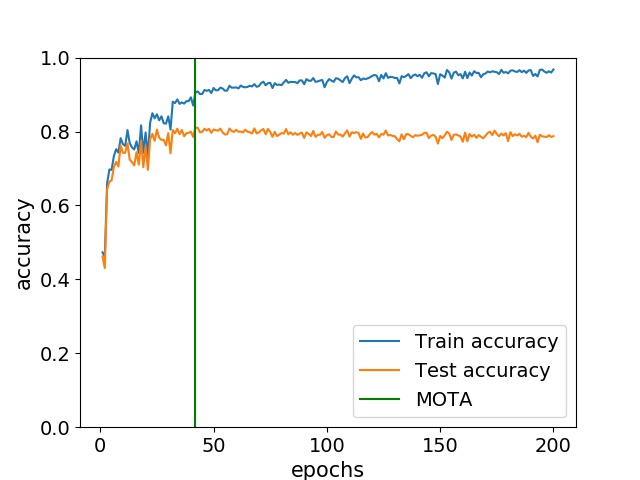}\\
  \caption{Label recall (left), Accuracy (right) for Feature-dependent noise on CIFAR-10 $\tau$ = 0.2 with ResNet32.}
  \label{fig: feature dependent noise}
  \vspace{-1\baselineskip}
\end{figure}

For example, the authors of \cite{Algan2020LabelLearning} synthetically generated feature-dependent noise, where the probability of mislabeling depends on the similarities in features of training examples. However, it can be observed in the left plot of Fig. \ref{fig: feature dependent noise} that the learning on noisy examples (orange line left plot) increases in the later stages of training, not satisfying the previously discussed learning behavior of feature-dependent label noise. Therefore, we create feature-dependent noisy datasets where the network significantly learns on both clean and noisy examples from the beginning of training. Our proposed approach also allows other researchers to easily create their own feature-dependent noisy datasets.

\section{Pseudo and Randomized Noisy Datasets}

\subsection{Motivation of Pseudo noise}

The key idea of developing Pseudo noisy datasets is inspired from the following human analogy. A person annotating a given dataset is likely to label an example correctly or incorrectly based on his proficiency (expert, trainee, etc.). For example, i) the labeler might be unable to distinguish among examples of different classes with similar features, ii) the labeler might find a specific class difficult to classify, etc. So, labelers with diverse expertise will annotate the same dataset with different percentages of noise rate and distributions.

Now, we relate the labeler to a DNN. The labels of a dataset annotated by a person with no prior knowledge is analogous to the predictions of a DNN with initialized weights. Furthermore, the DNN's weights at subsequent epochs can be comparable to expertise of different labelers. Therefore, we believe that the prior knowledge of a labeler can be equivalent to the learning distribution of the DNN on a training dataset (represented by the weights of the DNN). In other words, a labeler will annotate examples of a given dataset correctly or incorrectly based on their prior knowledge. Similarly, the expected prediction of the DNN on a new dataset will follow the distribution of the learning on the training dataset.

Based on this analysis, we can utilize the prediction errors of the DNNs on the training dataset to create noisy datasets. Thus, we train a DNN on a dataset with known ground truth and store the predictions of the network for each epoch. We further select the predictions at specific percentages of training accuracy to create Pseudo noisy datasets of desired noise rates. Therefore, the examples are mislabeled due to underlying learning misconception of DNN in Pseudo noisy datasets. Thus, we claim that the Pseudo noisy datasets would be closer to the feature-dependent label noise. Later, we will experimentally support this claim.

\subsection{Pseudo noise generation}

We select the predictions of the network on training examples at 1-$\tau$ \% training accuracy to create a Pseudo noisy dataset with a noise rate of $\tau$. However, it should be noted that, 1-$\tau$ \% training accuracy can be obtained at multiple epochs resulting in several Pseudo noisy datasets with the same noise rate $\tau$. Specifically, two factors vary in such noisy datasets with identical $\tau$: the, i) distribution of clean examples and ii) distribution of training examples, for each class. Thus, we define two parameters $\alpha$ and $\beta$, which measure the distribution of clean examples and training examples over all the classes respectively, to create Pseudo noisy datasets with identical $\tau$ and distinct noise distributions, as shown in Table \ref{tab:NTmatrix}. This can be related to creating different types of synthetic noisy datasets (symmetric and asymmetric) with identical $\tau$.

\begin{table}[t]
\caption{Two Noise distributions of 0.2 $\tau$.}
\begin{subtable}
    \centering
    % \small
    \addtolength{\tabcolsep}{-1pt}
    \begin{tabular}{|c|c|c|c|c|}
    \hline
    \textbf{$\hat{Y}$}&\multicolumn{4}{|c|}{Y}\\ 
    \hline
    &A&B&C&D\\
    \hline
    A&0.86&0.04&0.06&0.04\\
    B&0.1&0.77&0.07&0.06\\
    C&0.08&0.13&0.68&0.11\\
    D&0&0.06&0.1&0.84\\
    \hline
    
  \end{tabular}
  \end{subtable}
  \begin{subtable}
    \centering
    % \small
    % \resizebox{\columnwidth}{!}{%
    \begin{tabular}{|c|c|c|c|c|}
    \hline
    \textbf{$\Hat{Y}$}&\multicolumn{4}{|c|}{Y}\\ 
    \hline
    &A&B&C&D\\
    \hline
    A&0.81&0.11&0.04&0.04\\
    B&0.02&0.98&0&0\\
    C&0.32&0.09&0.53&0.06\\
    D&0.08&0.04&0&0.88\\
    \hline
    
   \end{tabular}
  \end{subtable}
  \label{tab:NTmatrix}
%   \vspace{-1\baselineskip}

\end{table}

%We broadly categorize Pseudo noisy datasets into two types: \emph{Close} (close to symmetric) shown in the left matrix Table \ref{tab:NTmatrix} and \emph{Medium} (right matrix). Medium type is more probable in real-world, as the noise distributions of human-error noisy datasets will usually not be symmetric.

Let $\hat{Y}$ and Y represent the true labels and noisy labels of training examples respectively in a Pseudo noisy dataset. Let $N_{ij}$ denote the noise distribution, where $i$ represents the true class and $j$ denotes the corresponding class in Y. $\alpha$ is calculated by taking the standard deviation ($\sigma$) of $N_{ij}$ at $i = j$, for all classes. Similarly, $\beta$ is found by calculating the $\sigma$ of distribution of training examples with reference to Y for each class, represented by $N_{j}$. In simple words, $N_{j}$ is obtained by calculating the column-wise sum of $N_{ij}$.

\vspace{-1\baselineskip}

\begin{align}
    \alpha = \sigma(diag(N_{ij})) &&  \beta = \sigma(N_{j})
\label{eq: 1}
\end{align}

In this work, we conduct experiments with shown $\alpha$ and $\beta$ values in Table \ref{tab:alphabeta}.

\subsection{Randomized noise generation}

To generate Randomized noisy datasets, examples are mislabeled by following a predetermined noise distribution obtained from Pseudo noisy datasets. Therefore, both Randomized and Pseudo noisy datasets have the exact same $N_{ij}$ (either left or right Table \ref{tab:NTmatrix}). However, the key difference is in the mislabeled examples. In Randomized noise, examples to be mislabeled are randomly picked, whereas in Pseudo noise, mislabeling of examples will occur due to underlying learning misconception of the DNN.

\section{Experiments}
\label{sec: experiments}

We train the network with the Adam optimizer, momentum of 0.9, batch size of 128 for 200 epochs with ResNet32 architecture. The initial learning rate is set to 0.001 and multiplied by 0.5, 0.25, 0.1 at 20, 30 and 40 epochs respectively for all our experiments. We create several Pseudo noisy datasets of varied noise rates $\tau$ from two clean benchmark datasets: CIFAR-10 and CIFAR-100.

\begin{table}[t]
\caption{alpha and beta values in our experiments.}
    \centering
    % \resizebox{\columnwidth}{!}{%

    \begin{tabular}{|p{2cm}|c|c|c|c|c|}
    \hline
    \textbf{Dataset}& \multicolumn{2}{|c|}{\textbf{CIFAR-10}} & \multicolumn{2}{|c|}{\textbf{CIFAR-100}}\\
    \hline
    \textbf{}&0.2&0.5&0.2&0.5\\
    \hline
    $\alpha$& 0.15 & 0.21 & 0.12 & 0.22 \\
    \hline
    $\beta$ & 0.31 & 0.4 & 0.29 & 0.78 \\ 
    \hline
    \end{tabular}
    % }
    \label{tab:alphabeta}
    \vspace{-1\baselineskip}

\end{table}

Metrics: Clean label recall at each epoch is calculated by dividing the total number of correctly predicted clean examples at that epoch over the total number of clean examples in the dataset. Likewise, noisy label recall is calculated. Let $LR_{clean}$ and $LR_{noisy}$ denote the vectors of clean and noisy label recall values respectively for all the epochs.

\subsection{Evaluation of Pseudo noise}

\begin{figure}[t]
\centering
%\begin{subfigure}
  \centering
  
%   \vspace{-0.3\baselineskip}
   \includegraphics[width= 0.255\textwidth]{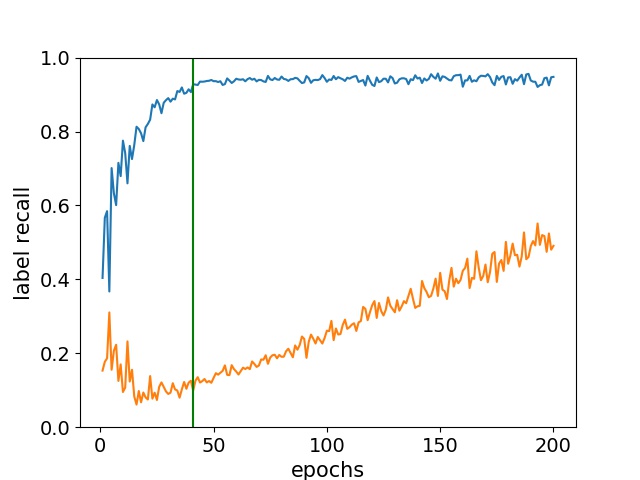}\includegraphics[width= 0.255\textwidth]{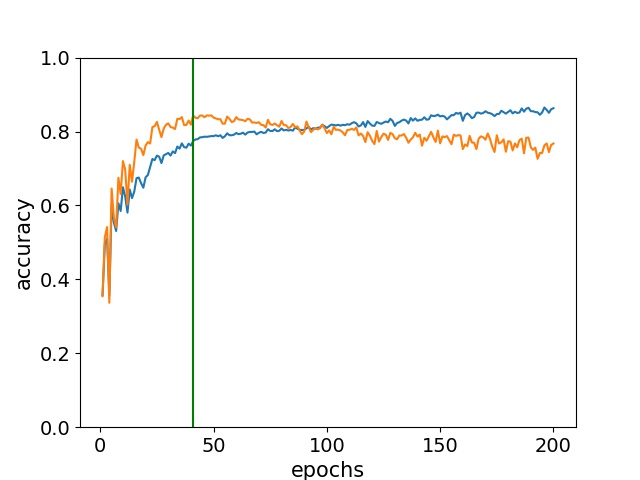}\\
    % \vspace{-0.3\baselineskip}
   \includegraphics[width= 0.255\textwidth]{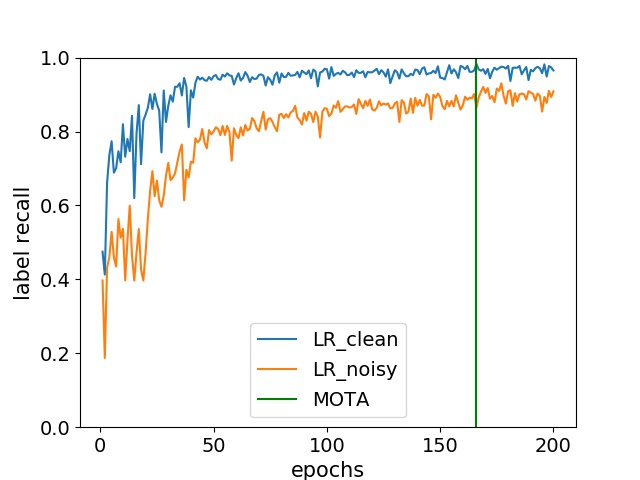}\includegraphics[width= 0.255\textwidth]{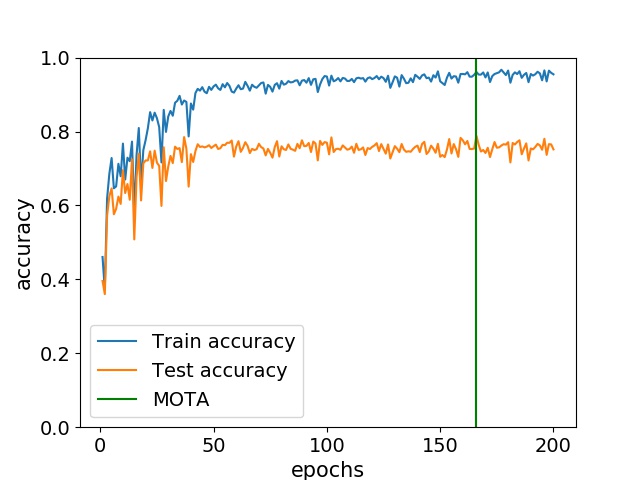}\\
   
  \caption{Label recall (left) and Accuracy (right) on CIFAR-10 with ResNet32 for $\tau$ = 0.2: Symmetric (top), Randomized (middle) and Pseudo (bottom).}
  \label{fig: LRacccomparison}
%\end{subfigure}
  \vspace{-1\baselineskip}
\end{figure}

We plot the label recall and accuracy of Randomized (top) and Pseudo (bottom) noise in Fig.\ref{fig: LRacccomparison}. It can be observed in the top left plot that $LR_{noisy}$ remains minimal in the initial stages of training similar to symmetric noise (Fig. \ref{fig: redmini}). As the $LR_{noisy}$ increases in the later stages, the test accuracy keeps dropping as shown in the top right plot.

\begin{figure}[t]
\centering
%\begin{subfigure}
  \centering
    \includegraphics[width= 0.255\textwidth]{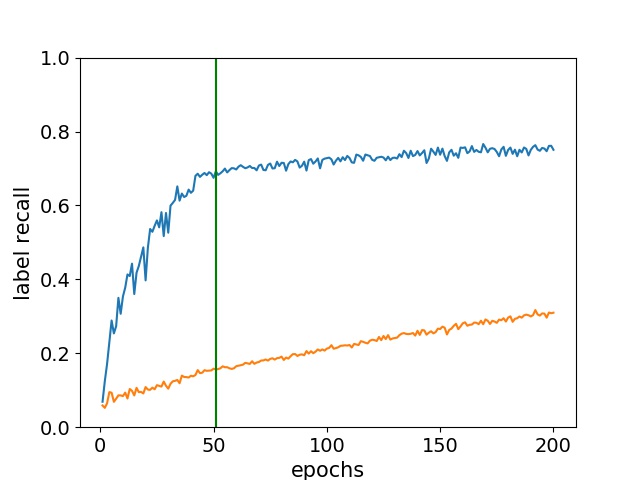}\includegraphics[width= 0.255\textwidth]{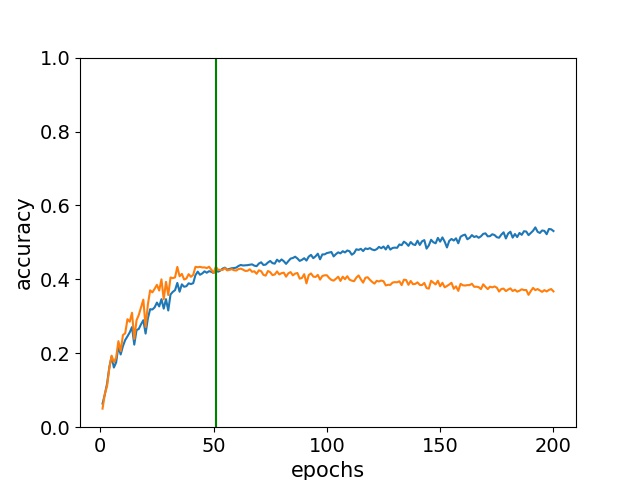}\\
    \vspace{-0.3\baselineskip}
    \includegraphics[width= 0.255\textwidth]{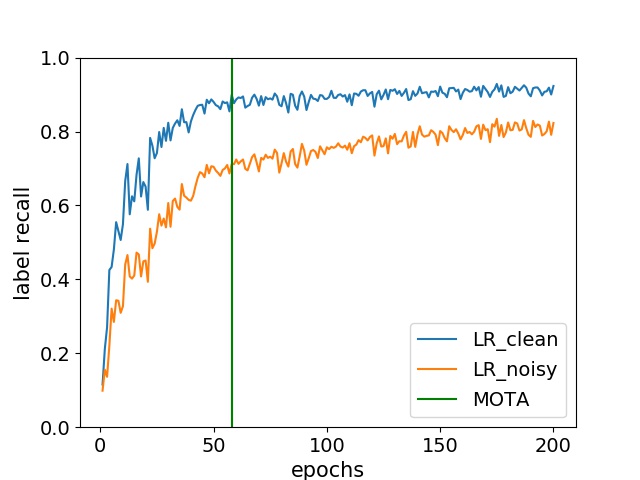}\includegraphics[width= 0.255\textwidth]{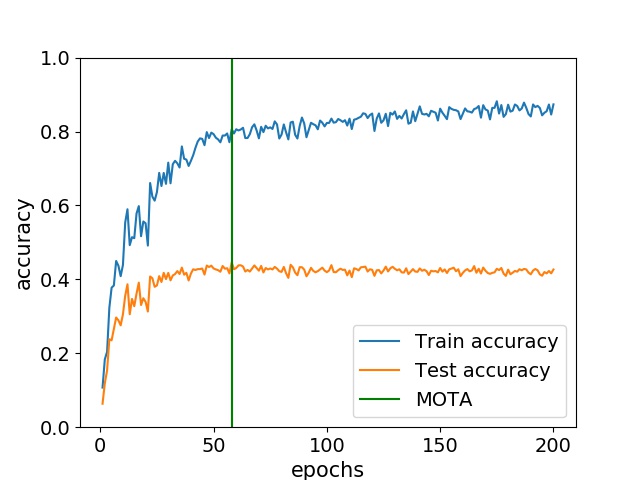}\\
    \caption{Label recall (left) and Accuracy (right) on CIFAR-100 with ResNet32 for $\tau$ = 0.5: Randomized (top) and Pseudo (bottom).}
  \label{fig: LRacccifar100}
%\end{subfigure}
  \vspace{-1\baselineskip}
\end{figure}
Pseudo noise on the other hand exhibits a different learning behavior as can be seen in the bottom plots. It can be observed that the network fits both clean ($LR_{clean}$) and noisy ($LR_{noisy}$) examples from the beginning of the training at a higher rate with Pseudo noise. Consequently, it can be noticed that the test accuracy (bottom right) does not vary much in the later stages of training. This clearly satisfies the previously discussed learning behavior of feature-dependent label noise.

It should be noted that the $\tau$ and noise distribution in the top and bottom plots are exactly the same. The only difference is that the examples are randomly mislabeled in Randomized noise, whereas examples are mislabeled due to underlying misconception of DNN in Pseudo noise.

\begin{figure}[t]
\centering
%\begin{subfigure}
  \centering
    \includegraphics[width= 0.255\textwidth]{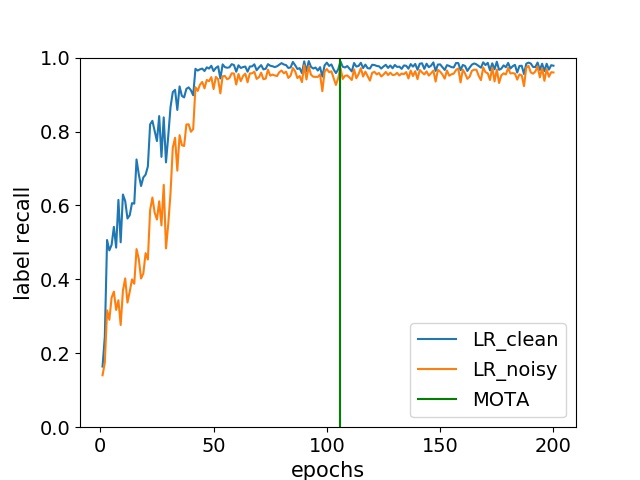}\includegraphics[width= 0.255\textwidth]{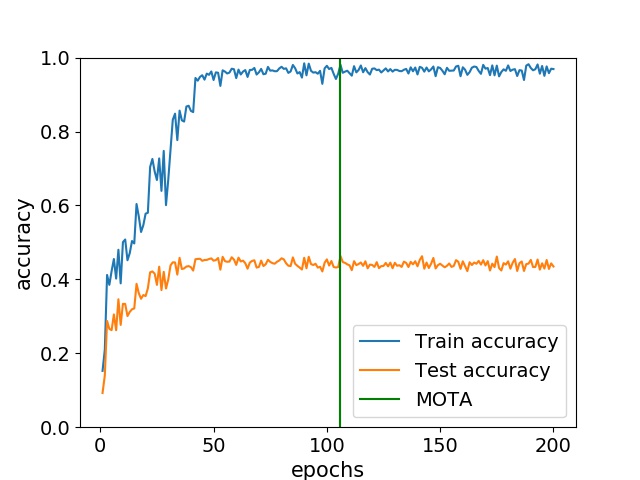}\\
    \vspace{-0.3\baselineskip}
    \includegraphics[width= 0.255\textwidth]{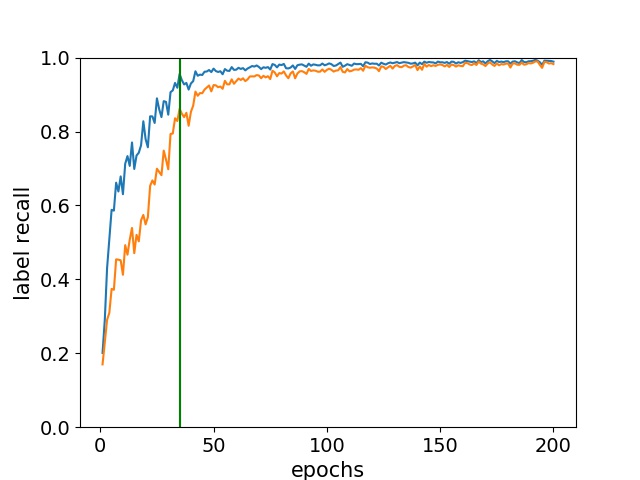}\includegraphics[width= 0.255\textwidth]{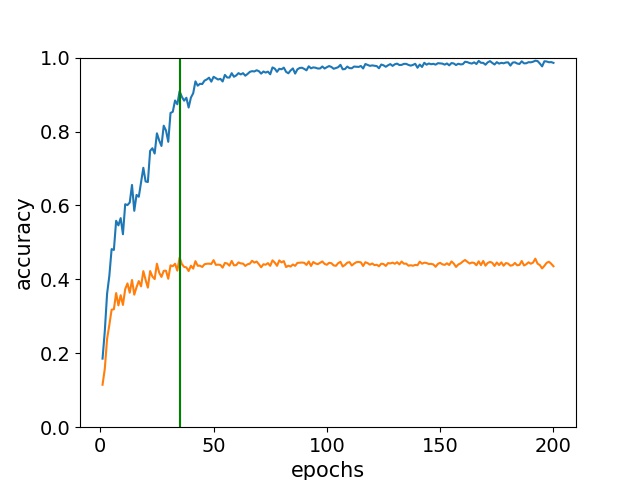}\\
    \caption{Label recall (left) and Accuracy (right) on CIFAR-100 for $\tau$ = 0.5 Pseudo: CNN9 (top) and ResNet110 (bottom).}
  \label{fig: LRacccifar100architect}
%\end{subfigure}
  \vspace{-1\baselineskip}
\end{figure}

\begin{figure}[t]
\centering
%\begin{subfigure}
  \centering
    \includegraphics[width= 0.255\textwidth]{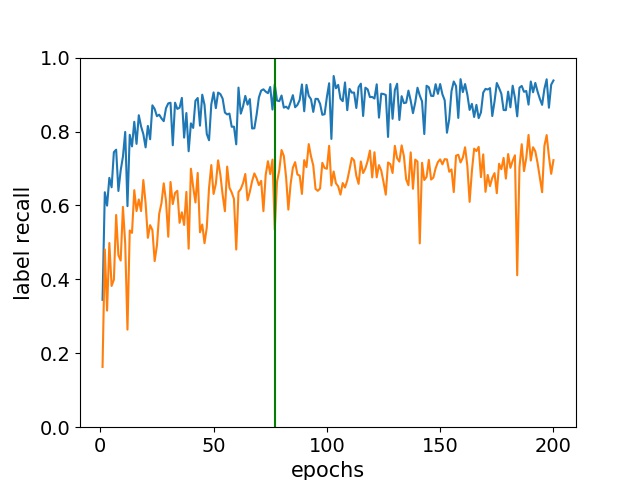}\includegraphics[width= 0.255\textwidth]{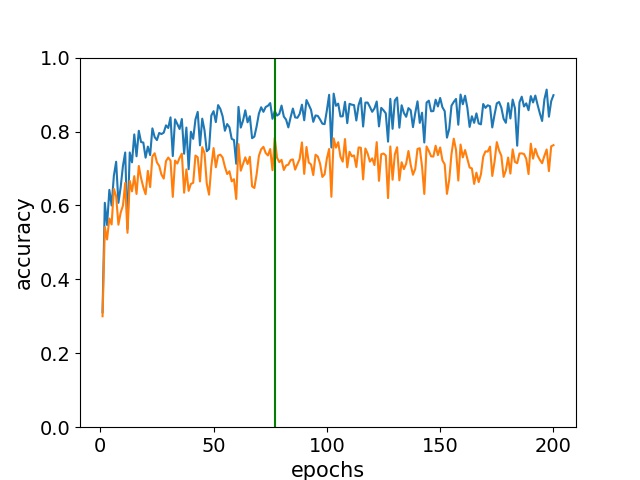}\\
    \vspace{-0.3\baselineskip}
    \includegraphics[width= 0.255\textwidth]{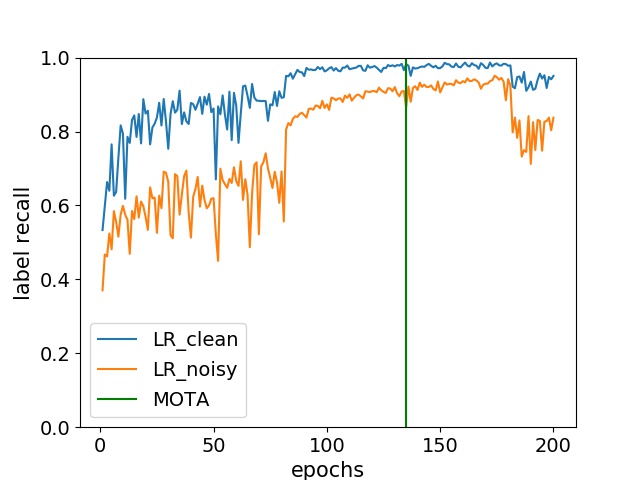}\includegraphics[width= 0.255\textwidth]{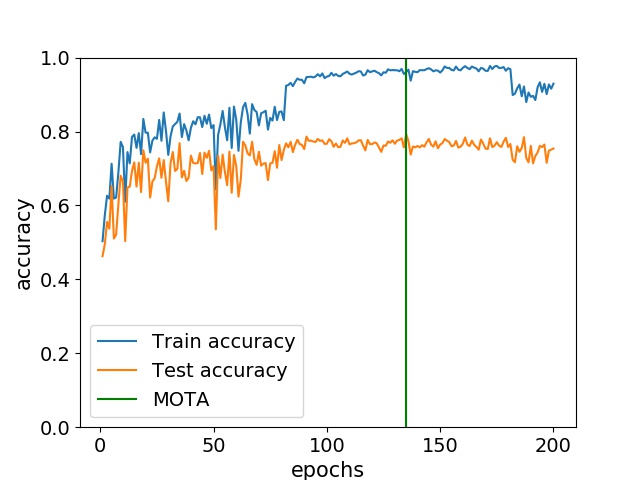}\\
    \caption{Label recall (left) and Accuracy (right) on CIFAR-10 for $\tau$ = 0.2 Pseudo with ResNet32: constant learning rate (top) and decay 2 (bottom).}
  \label{fig: LRacclearnrate}
%\end{subfigure}
  \vspace{-1\baselineskip}
\end{figure}

The above discussed learning behavior of DNNs on Randomized and Pseudo noise is consistent across different conditions: i) higher $\tau$ and harder dataset (Fig. \ref{fig: LRacccifar100}), ii) different architectures (Fig. \ref{fig: LRacccifar100architect}) and iii) varying learning rate (Fig. \ref{fig: LRacclearnrate}). The top plot of Fig. \ref{fig: LRacclearnrate} is with constant initial learning rate (0.001), while the bottom plot is obtained when the initial learning rate is multiplied by 0.1 at 80, 120, 160 epochs and 0.5 at $180^{th}$ epoch (referred as decay 2 in figure). Note that the drop towards the end is due to the change in learning rate. These results clearly indicates that Pseudo noisy datasets resemble feature-dependent noisy datasets.

\subsection{Validity of Pseudo noise}

In the plots (Fig \ref{fig: LRacccomparison}-\ref{fig: LRacccifar100}), we previously noticed that the learning behavior of Randomized noisy datasets doesn't resemble feature-dependent noise. This suggests that the noisy datasets created with \emph{predetermined noise distributions} don't mimic the feature-dependent label noise, even when the noise distribution is same as Pseudo noise.

\begin{figure}[t]
  \centering
  \includegraphics[width=0.255\textwidth]{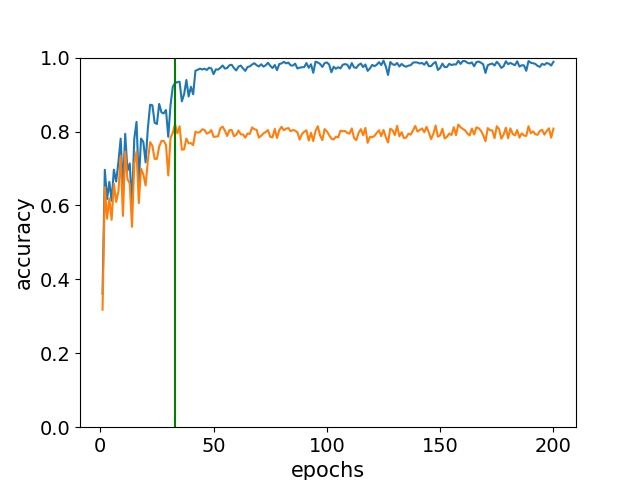}\\
  \caption{Accuracy with ResNet32: CIFAR-10 for $\tau$ = 0.18 Pseudo (left).}
  \label{fig: close}
  \vspace{-1\baselineskip}
\end{figure}

On the other hand, we verify the learning behavior of Pseudo noise when the noise distribution is similar to \emph{symmetric} noise. It can be observed in Fig. \ref{fig: close} ($\alpha$ = 0.11, $\beta$ = 0.15) that the network still learns on both clean and noisy examples from the beginning of training. This clearly shows that irrespective of the noise distribution, Pseudo noisy datasets are feature-dependent noisy datasets.

\subsection{Learning is easier with feature-dependent label noise}

We further utilize the learning behavior of DNNs to support our hypothesis that learning is easier with feature-dependent label noise compared to Random noise.

It can be observed in the top plots of Fig.\ref{fig: LRacccifar100} that the training accuracy (top right blue line) of Randomized noise for harder dataset CIFAR-100 and higher noise rate $\tau$ of $0.5$ is very low, as the network needs to learn harder patterns to find relationship among the clean examples and randomly mislabeled examples. As a result, the network fits mostly clean examples in the initial stages.

On the other hand, the right bottom plot of Fig.\ref{fig: LRacccifar100} clearly show that the training accuracy is higher for Pseudo noise despite harder dataset and higher noise rate. This indicates that the network can easily fit both clean and noisy examples by learning easier patterns from the beginning of training, as the mislabeling is caused due to underlying misconception of the DNN. Therefore, this proves that learning with Pseudo noisy datasets (feature-dependent label noise) is much easier compared to random noisy datasets.

\subsection{Comparison of existing noise-robust training methods}

In this section, we compare several existing training methods on Pseudo and Randomized noise with CIFAR-10 and CIFAR-100 on different $\tau$ values as shown in Table \ref{tab:comparisonacc}. In particular, we compare the following broad approaches: i) sample selection methods (Co-teaching \cite{Han2018Co-teaching:Labels} and INCV \cite{Chen2019UnderstandingLabels}), ii) loss correction methods (F-correction \cite{Patrini2017MakingApproach}, Bootsoft \cite{Reed2015TrainingBootstrapping} and Joint \cite{Tanaka2018JointLabels}), iii) robust loss functions (SCE \cite{Wang2019SymmetricLabels} and GCE\cite{Zhang2018GeneralizedLabels}), iv) early stopping (AutoTSP\cite{Kamabattula2020IdentifyingData} and NHA \cite{Song2019Prestopping:Noise}) and v) standard training without early stopping (Standard).

We note the test accuracy and label recall at two points: MOTA (where maximum test accuracy is obtained) and final epoch (referred as MOTA/ final epoch in the Table) for all the approaches, except the early stopping methods (accuracy at stop point is reported).

\begin{table*}[t]
\caption{Average Test accuracy (\%, 3 runs) comparison of Randomized and Pseudo noise with ResNet32 (maximum/final test accuracy).}
    \centering
    \resizebox{\linewidth}{!}{%

    \begin{tabular}{|p{2cm}|c|c|c|c|c|c|c|c|}
    \hline
    \textbf{Dataset}& \multicolumn{4}{|c|}{\textbf{CIFAR-10}} & \multicolumn{4}{|c|}{\textbf{CIFAR-100}}\\[2pt]
    \hline
    \textbf{Noise rate}& \multicolumn{2}{|c|}{0.2} & \multicolumn{2}{|c|}{0.5}& \multicolumn{2}{|c|}{0.2}& \multicolumn{2}{|c|}{0.5}\\[2pt]
    \hline
    \textbf{Noise type}&Randomized&Pseudo&Randomized&Pseudo&Randomized&Pseudo&Randomized&Pseudo\\[2pt]
    \hline
    \hline
    Standard& 84.37/76.75 & 78.85/75.17 & 70.13/56.74 &52.49/50.99 & 57.63/52.13 & 58.71/56.75 & 43.4/36.26 & 45.23/44.35 \\[2pt]
    \hline
    Co-teaching \cite{Han2018Co-teaching:Labels}& 88.45/\textbf{88.01} & 80.1/\textbf{79.07} & 67.26/67.11 & 51.1/50.32 & 62.41/\textbf{62.08} & 61.08/\textbf{60.97} & 45.4/\textbf{44.85} & 44/43.59 \\[2pt]
    \hline
    INCV \cite{Chen2019UnderstandingLabels} & 88.13/87.71 & 78.78/77.69 & 68.4/\textbf{68.3} & 52.25/51.22 & 58.29/58.07 & 60.25/60 &40.12/40& 44.4/43.57 \\ [2pt]
    \hline
    F-correction \cite{Patrini2017MakingApproach}& 84.72/76.7 & 78.35/74.67 & 68.82/54.85 & 53.62/51.06 & 56.95/53.02 & 58.73/56.76 & 43.72/35.39 & 44.6/42.9 \\[2pt]
    \hline
    SCE \cite{Wang2019SymmetricLabels}  & 83.78/82.32 & 78.22/73.79 & 66.02/64.77 & 52.14/50.99 & 54.6/53.41 & 55.15/54.53 & 40.37/38.44 & 42.67/41.92  \\[2pt]
    \hline
    GCE \cite{Zhang2018GeneralizedLabels}   & - & 79.57/74.81 & - & 52.56/50.23 & - & 57.04/55.12 & 42.26/40.6 & 44.55/41.38  \\[2pt]
    \hline
    Joint \cite{Tanaka2018JointLabels}  & 86.7/80.96 & 79.01/74.01 & 78.56/68.11 & 53.43/51.37 & 57.33/54.25 & 58.59/57.73 & 49.13/44.18 & 46.77/43.96  \\[2pt]
    \hline
    Bootsoft \cite{Reed2015TrainingBootstrapping}   & 84.46/75.41 & 78.83/77.16 & 68.34/57.84 & 52.52/51.31 & 57.96/52.13 & 58.71/56.71 & 44/38.24 & 44.89/43.73  \\[2pt]
    \hline
    AutoTSP \cite{Kamabattula2020IdentifyingData}& 81.21 & 75.51 & 63.32 & \textbf{52.49} & 55.73 & 56.44 & 43.29 & \textbf{44.06} \\[2pt]
    \hline
    NHA \cite{Song2019Prestopping:Noise}& 81.07 & 74 & 64.61 & 42.24 & 52.13 & 56.3 & 39.33 & 35.57 \\[2pt]
    \hline
    %Train on Clean & 88 & 85.77 & 84.32 & 73.07 & 60.5 & 60.66 & 53.88 & 51.58\\ 
    %\hline
    \end{tabular}
    }
    \label{tab:comparisonacc}
    % \vspace{-1\baselineskip}

\end{table*}

\begin{table*}[t]
\caption{Noisy Label Recall of randomized and pseudo noise at Maximum/ Final test accuracy with ResNet32.}
    \centering
    %\begin{adjustbox}
    \resizebox{\linewidth}{!}{%
    \Huge
    \renewcommand{\arraystretch}{1.25}
    \begin{tabular}{|c|c|c|c|c|c|c|c|c|}
    \hline
    \textbf{Dataset}& \multicolumn{4}{|c|}{\textbf{CIFAR-10}} & \multicolumn{4}{|c|}{\textbf{CIFAR-100}}\\
    \hline
    \textbf{Noise rate}& \multicolumn{2}{|c|}{0.2} & \multicolumn{2}{|c|}{0.5}& \multicolumn{2}{|c|}{0.2}& \multicolumn{2}{|c|}{0.5}\\
    \hline
    \textbf{Noise type}&Randomized&Pseudo&Randomized&Pseudo&Randomized&Pseudo&Randomized&Pseudo\\
    \hline
    \hline
    Standard& 0.12/0.49 & 0.78/0.9 & 0.14/0.45 & 0.76/0.84 & 0.11/0.32 & 0.53/0.76 & 0.14/0.3 & 0.65/0.82 \\
    \hline
    Co-teaching \cite{Han2018Co-teaching:Labels}& 0.09/0.12 & 0.48/0.67 & 0.13/0.14 & 0.46/0.61 & 0.07/0.19 & 0.31/0.62 & 0.09/0.2 & 0.31/0.54 \\
    \hline
    SCE \cite{Wang2019SymmetricLabels}  & 0.09/0.16 & 0.65/0.74 & 0.13/0.19 & 0.71/0.75 & 0.07/0.13 & 0.34/0.51 & 0.13/0.22 &  0.54/0.68   \\
    \hline
      
    \end{tabular}
   }
   %\end{adjustbox}
    \label{tab:comparisonlrnoisy}
    \vspace{-1\baselineskip}
  
\end{table*}

As discussed earlier, the MOTA and the end test accuracy of Standard method drastically vary for Randomized noise, while the variance is small for Pseudo noise. It can also be observed that the MOTA obtained for Randomized noise is much higher than the Pseudo for CIFAR-10 dataset, because higher noisy label recall is obtained for Pseudo compared to Randomized noise as shown in Table \ref{tab:comparisonlrnoisy}. However, the maximum test accuracy is similar for both Pseudo and Randomized with CIFAR-100. It is likely because, generalization becomes harder for higher number of classes.

Among the existing training methods that we compared, it can be observed that the sample selection methods outperform all the others on both Pseudo and Randomized noise. Specifically, co-teaching consistently achieves higher test accuracy across different noise ratios and datasets, while INCV becomes inaccurate with higher number of classes. It can be further observed that the generalization performance is not improved than Standard with the remaining approaches for both Pseudo and Randomized noise. It is likely because, the methods are sensitive to small perturbations in their assumed noise distributions. Among these remaining methods, i) the test accuracy of SCE remains consistent from MOTA till the end of training across different cases, while it keeps dropping for other methods much like Standard training and ii) AutoTSP achieves higher accuracy at the end of training for 0.5 $\tau$ on CIFAR-10 and CIFAR-100 than the other approaches, with just finding a training stop point on Standard training, unlike the remaining approaches which modify the training framework.

As previously mentioned, co-teaching achieves higher test accuracy than the Standard even for Pseudo noise. However, the $LR_{noisy}$ in Table \ref{tab:comparisonlrnoisy} implies that co-teaching still significantly learns on noisy examples with Pseudo noise, while it remains minimal with Randomized noise.
This indicates that the current training methods are not generalizable to feature-dependent noisy datasets. 

These results altogether substantiate the significance of Pseudo noise. We believe that our proposed Pseudo noisy datasets together with the ground truth information can be utilized to study and develop robust training methods with feature-dependent label noise.

% \section{Discussion}

% We discuss about the validity of early stopping. The authors of \cite{Li2019GradientNetworks} show that it is desirable to stop training in the early stages, while \cite{Jiang2020BeyondLabels} suggests otherwise. On the other hand, our results (green vertical line in Fig. \ref{fig: LRacccomparison}-\ref{fig: close}) indicate that early stopping achieves better generalization performance with random noise and, it's significance is reduced with complex label noise. So, this supports the claim of \cite{Algan2020LabelLearning} that the label noise problem needs to be treated separately. Therefore, in order to effectively train with noisy labeled datasets, we believe that the type of label noise needs to be first understood.

\section{Conclusion}
 
In this work, we proposed Pseudo noisy datasets by utilizing the predictions of DNNs that also provide ground truth to effectively study feature-dependent label noise. We first established through the learning behavior of DNNs that Pseudo noisy datasets mimic feature-dependent label noise across different conditions, where distinguishing between clean and noisy examples is harder. We further generated a Randomized noise with the same noise distribution as that of Pseudo noise and empirically showed that i) learning is easier with feature-dependent label noise compared to random noise, ii) regardless of the noise distribution, pseudo noisy datasets resemble feature-dependent label noise and iii) several existing noise-robust training methods are not generalizable to feature-dependent (Pseudo) label noise. Therefore, we believe that our proposed approach will allow other researchers to create their own feature-dependent noisy datasets effortlessly in various domains in order to develop effective training methods.

\bibliographystyle{IEEEtran}
\bibliography{references}

\end{document}